\documentclass{acm_proc_article-sp}

\usepackage{graphicx}
\usepackage{epsfig}

\usepackage{color}
\usepackage{booktabs}

\usepackage[english]{babel}
\usepackage[utf8]{inputenc}
\usepackage[colorinlistoftodos]{todonotes}

\usepackage{amsmath}    
\usepackage{mathtools}  
\usepackage{amssymb}    
\usepackage{verbatim}   
\usepackage{color}      
\usepackage{hyperref}   
\usepackage{multicol}  %
\usepackage{multirow}
\usepackage{ifthen}

\usepackage{graphicx}
\usepackage{caption}
\usepackage{subcaption}

\usepackage{lipsum}       
\usepackage{changepage}   
\usepackage{enumitem}

\usepackage{soul}

\begin{document}

\conferenceinfo{}{Bloomberg Data for Good Exchange 2016, NY, USA}

\title{Flint Water Crisis: Data-Driven Risk Assessment Via Residential Water Testing}

\numberofauthors{12}
\author{
\alignauthor
Jacob Abernethy\\
       \email{jabernet@umich.edu}
\alignauthor
Cyrus Anderson\\  
	\email{andersct@umich.edu}
\alignauthor
Chengyu Dai\\
       \email{daich@umich.edu}
\and
\alignauthor
Arya Farahi\\
       \email{aryaf@umich.edu}
\alignauthor
Linh Nguyen\\
       \email{lvnguyen@umich.edu}
\alignauthor
Adam Rauh\\
       \email{amrauh@umich.edu}
\and
\alignauthor 
Eric Schwartz\\
       \email{ericmsch@umich.edu}
\alignauthor
Wenbo Shen\\
       \email{shenwb@umich.edu}
\alignauthor
Guangsha Shi\\
       \email{guangsha@umich.edu}
\and
\alignauthor
Jonathan Stroud\\
       \email{stroud@umich.edu}
\alignauthor 
Xinyu Tan\\
       \email{xinyutan@umich.edu}
\alignauthor
Jared Webb\\
       \email{jaredaw@umich.edu}
\and
\alignauthor
Sheng Yang\\
       \email{physheng@umich.edu}
\begin{center}
\affaddr{University of Michigan}\\
\affaddr{Ann Arbor, MI}
\end{center}
}

\maketitle

\begin{abstract}
Recovery from the Flint Water Crisis has been hindered by uncertainty in both the water testing process and the causes of contamination. In this work, we develop an ensemble of predictive models to assess the risk of lead contamination in individual homes and neighborhoods. To train these models, we utilize a wide range of data sources, including voluntary residential water tests, historical records, and city infrastructure data.  Additionally, we use our models to identify the most prominent factors that contribute to a high risk of lead contamination. In this analysis, we find that lead service lines are not the only factor that is predictive of the risk of lead contamination of water. These results could be used to guide the long-term recovery efforts in Flint, minimize the immediate damages, and improve resource-allocation decisions for similar water infrastructure crises. 
\end{abstract}





\section{Introduction}

The Flint Water Crisis began in April 2014 when the city of Flint, Michigan switched its water supply from Lake Huron to the Flint River as a temporary cost-saving measure. Not long afterwards, the water in many Flint residences was found to be contaminated with dangerously high levels of lead. It was discovered that the highly-corrosive water drawn in from the Flint River was not treated with the proper anti-corrosive chemicals prior to the switch, causing lead particles to leech into the water supply from the lead pipes that comprise much of the city's aging infrastructure. In many places, the levels of lead in the water exceeded one hundred times the federal actionable level of 15 parts per billion (ppb), and blood-borne lead levels in children increased noticeably since the switch \cite{hanna:2016}. Since lead-contaminated water poses significant health risks, particularly for children \cite{Toronto2014}, the mayor of Flint declared the city to be in a state of emergency in December 2015. Flint has since returned to the water shipped in from Lake Huron. However, lead contamination has yet to return to safe levels. The city of Flint now faces a daunting infrastructural problem: find which homes are most drastically affected by lead contamination and repair their plumbing systems. Conventional wisdom says that homes with lead service lines are at the highest risk of contamination, and it is estimated that Flint has over 8000 such service lines \footnote{http://www.freep.com/story/news/2016/02/22/um-study-more-than-8000-lead-service-lines-flint/80750870/}. This problem has gathered significant national attention, and the city of Flint is now under pressure to repair the infrastructural issues as quickly and efficiently as possible.

Although many believe that repairing the lead service lines will remove lead contamination, there are a number of difficulties that complicate this proposed solution. First, it is not clear whether repairing all of Flint's lead service lines will eliminate the problem. Even if the contact between corrosive Flint River water and lead service lines was the original cause of lead contamination, other regions in the water delivery pipeline now contain contaminated water. This water may remain in the infrastructure if the surrounding water systems are unused for prolonged periods, and may continue to contaminate nearby residences when fluctuations in water pressure eventually push it through the system. Flint is particularly at high risk of this phenomenon because it has one of the highest property vacancy rates in the country. This is just one of many factors besides lead service lines that may put a particular home at risk. Understanding the complex relationship between the contamination problem and features of the infrastructure will allow policymakers to more effectively target the problem and minimize damage.

The second complication is that it is not clear which homes are most effected by contamination. To address this, the city has implemented a voluntary residential water testing program that allows residents to collect their own water and submit it for testing at a local center. However, not all homes have been tested, and the lead levels in individual homes fluctuate frequently, making it difficult to obtain accurate measurements. If an accurate risk assessment can be made for each home, infrastructural updates can be prioritized to the places they are most needed.

In this work, we take a data-driven approach to aid policymakers with these issues. Specifically, we make two contributions. First, we develop predictive models that predict which homes are at the highest risk of containing dangerous lead levels in their water. Second, we analyze these models to see which features are most predictive of lead contamination. This indicates which  risk factors should be addressed first when repairing the infrastructure. Our predictive models give the estimated probability that a home has a water lead level above the federal action level of 15 ppb. These models are then composed into an ensemble which  more accurately predicts these probability estimates than any particular model on its own. We hope that our predictive models and analyses will prove useful to policymakers and the people of Flint, and help guide the decision making process to mitigate the damage done by this crisis. We also hope our approach will be applicable to the other urban areas around the country with aging infrastructures and lead contamination problems.

\subsection*{Related Work}

Several prior works have taken data-driven approaches to solving infrastructural problems. Notably, \cite{lau:2015} use Support Vector Machines to predict the risk of fires in residences, and \cite{potash2015predictive} use similar predictive models to predict lead poisoning in children. Most related to our work is \cite{li:2014}, who use hierarchical beta processes to predict pipe failures in the water system of Sydney, Australia.

Regarding the Flint Water Crisis, much of the work up until this point has been conducted by Marc Edwards' team from Virginia Tech \footnote{http://www.flintwaterstudy.org}. Their efforts have helped raise awareness and reveal the severity of the problem. In addition, \cite{baum:2016} provides an overview of the water crisis and discusses strategies for risk management in Flint. To the best of our knowledge, we are the first to apply predictive modeling techniques to aid with the Flint Water Crisis.

\section{Data}

Our predictive models incorporate a diverse range of datasets from the city of Flint. Much of this data is publicly available from the state of Michigan, and other components were provided by the city at our request, as noted.

\subsection{Residential Lead Tests}

\begin{figure}[h]
\centering
\vspace{0.1in}
\includegraphics[width=0.5\textwidth]{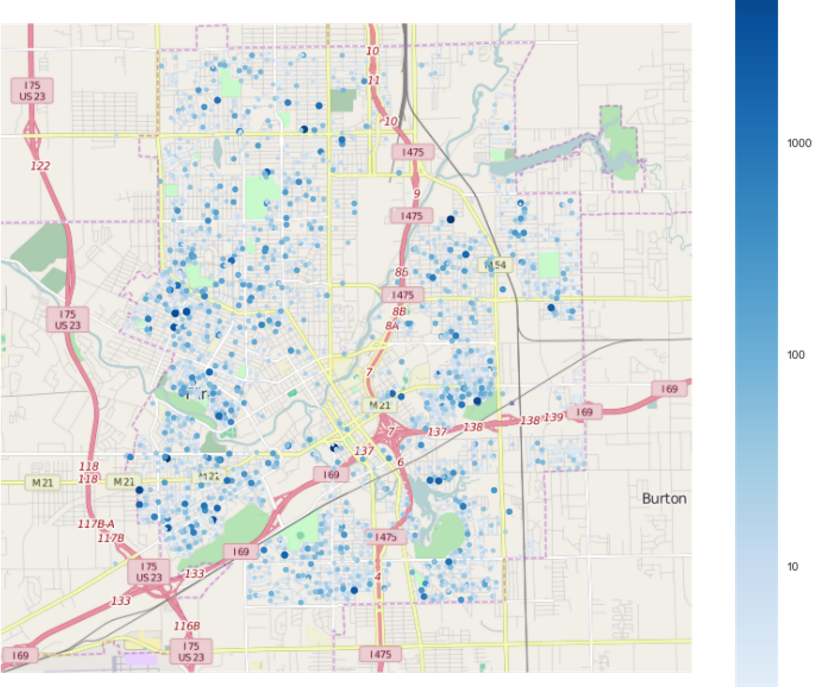}
\caption{Locations of voluntary residential water tests in Flint. Color corresponds to the level of lead contamination (parts per billion). We observe that elevated lead readings are highly geographically diverse.}
\label{fig:heatmap}
\end{figure}

Most of the lead sampling data from Flint comes from water samples submitted voluntarily by residents. The city of Flint provides free water testing services to all of its residents, who are able to pick up testing kits from a local distribution center.  They then collect water from their own homes, and submit the samples to be analyzed by the Michigan Department of Environmental Quality. Since this program began in September 2015, over 15,400 tests have been conducted, and the results have been made available online \footnote{http://www.michigan.gov/flintwater/}. For each submitted sample, we are given the date the sample was submitted, the lead and copper levels, and the address of the residence. In Figure ~\ref{fig:heatmap}, we show the locations and lead readings for these tests.

\subsection{Parcel data}

The city provided us with detailed records of the 55,893 parcels of land in Flint. This data contains information on the property's age, location, and value, in addition to dozens of other property characteristics. This data is not publicly available online.

We match the address of each residential lead sample to a parcel of land within the city. Those that did not correspond to Flint parcels were discarded. Because a parcel can contain multiple residences, and because residents are free to submit as many tests as they would like, we often have multiple tests that correspond to a single parcel. On the other hand, because many properties in Flint are vacant and residents are not required to submit tests, many parcels have no associated lead test.

\subsection{Service line data} \label{sec:serviceLines}

\begin{figure}
\centering
\includegraphics[width=0.48\textwidth]{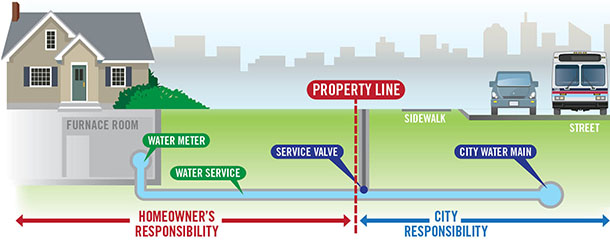}
\caption[Caption for LOF]{\label{fig:slschmatic}
Service lines connect water main in street to homes. Private and public portions of service lines are split at property line.}
\end{figure}

As the Water Crisis was brought into full view and officials began to realize the severity of the situation, a search began to determine the primary culprits; that is, the location of lead metals that were producing the elevated lead water readings. Within days of the media blitz, a large part of the discussion turned to the \emph{water service lines}, the pipes that connect each property in Flint to the water distribution system, often called the ``water main.'' Service lines can be made out of any number of materials, including copper, galvanized steel, plastic, lead, as well as various metal alloys. 

A home's water service line is typically composed of two different segments: the \emph{public} service line which is the pipe connecting the water main to the property ``curb box,'' an underground device owned by the municipality that contains a shutoff valve; and the \emph{private} service line, which connects the curb box through front lawn, and usually runs into the basement and attaches to the home's water meter. See Figure \ref{fig:slschmatic} for schematic\footnote{http://www.calgary.ca/UEP/Water/PublishingImages/\\Water\_Service\_Property\_Line\_610px.jpg. Image copyright City of Calgary}.

\begin{table}
\centering
\begin{tabular}{lr}
\toprule
\textbf{SL Record} & \textbf{$\#$ Parcels} \\
\midrule
Copper           &    25843 \\
Unkown/Other     &    13090 \\
Galvanized/Other &    12261 \\
Copper/Lead      &     4161 \\
Copper/?         &      149 \\
Tubeloy          &      118 \\
Lead             &      111 \\
Lead/Tubeloy     &       59 \\
\bottomrule
\end{tabular}
\caption{Summary table of service line materials according to city records. There are many difficult to interpret labels, including the use of `?' as well as pairs of metal names, without any explanation of whether these pairs correspond to public/private, or private/public.}\label{tab:serviceLines}
\end{table}

\begin{figure}[h]
\centering
\includegraphics[width=5cm]{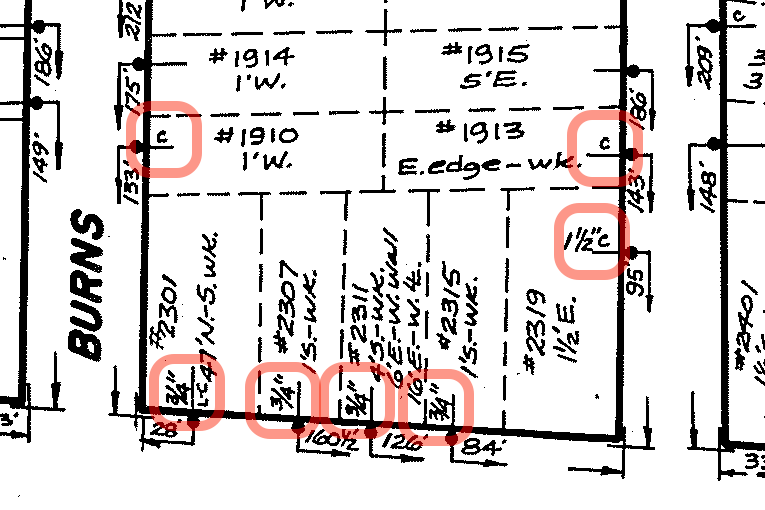}
\caption{Service Line Records. When Flint's water troubles began, the city was unable to produce accurate records of the service line materials for all properties. They were eventually able to find a set of annotated maps with various markings representing the material used in the pipes (circled in red). \textbf{C} stands for Copper, \textbf{L-C} stands for Lead/Copper, and \textbf{3/4"} stands for Galvanized. A large fraction of the records were blank.}
\label{fig:leadservicelines}
\end{figure}

Many municipalities have very accurate and updated records describing service line attributes (material, length, location, etc.). The City of Flint, on the other hand, initially struggled to produce any records at all. Ultimately they discovered a set of 45,000 $3"\times 5"$ index cards, as well as a set of municipal maps from the water department\footnote{http://www.npr.org/2016/02/01/465150617/flint-begins-the-long-process-of-fixing-its-water-problem} with handwritten annotations. A sample of these maps is found in Figure~\ref{fig:leadservicelines}. The information in these maps was painstakingly digitized by a group of undergraduate students at the University of Michigan, Flint, within the GIS center. This project was spearheaded by Dr. Marty Kaufman, a researcher and the director of the center. Table \ref{tab:serviceLines} summarize service line materials according to city records.

Some entries in the service line material field have the form ``X/Y", for example ``Copper/Lead". These documents do not spell out exactly what was intended by these duplicate labels, but our evidence suggests that \emph{typically} this implies that the second label (``Lead'' in ``Copper/Lead'') describes the public service line material, whereas the first label describes the private service line. An entry that is simply given as ``Copper" may refer to both sections or either. Lastly, there are a number of entries in the records that say ``Copper/?" for the service line material, indicating unknown markings for the service line on the original handwritten records. Many other records are simply blank.

Since the crisis began, the Michigan DEQ has solicited plumbers and others to take part in a large number of in-home inspections, now totaling over 3300 parcels. For this set of homes we have verified results on the type of material in the \emph{private portion} of the service line. Thus we are able to estimate the accuracy of the city records for the private part of the service line. We report these results in Table~\ref{tbl:slrecords}.

\begin{table}
\begin{tabular}{lrrr}
\toprule
& \multicolumn{3}{c}{Private SL Material (via Inspection)} \\
City Record &  \textbf{Copper} &  \textbf{Galvanized} &  \textbf{Lead} \\
\midrule
Copper           &    1535 &          38 &    13 \\
Copper/Lead      &     685 &          58 &    40 \\
Galvanized/Other &     177 &         237 &     4 \\
Lead             &       7 &           4 &    24 \\
Tubeloy          &      28 &           1 &     2 \\
Unkown/Other     &     302 &         279 &    40 \\
\bottomrule
\end{tabular}
\caption{The ``confusion matrix'' for the city records on service line material versus what was discovered upon inspection; inspections only determined the private portion of the service line. Every row in this table corresponds to a label in the city records, whereas every column corresponds to the result of an inspection. For example, we report that there are 1,535 homes where the city had a record of copper service lines which were confirmed by an inspection, yet there were 13 such homes where lead was found upon inspection.} \label{tbl:slrecords}
\end{table}

\subsection{Fire hydrant data}

   It was hypothesized by the City of Flint planning department that the locations and types of the many fire hydrants scattered around the city would be helpful in understanding the city's water infrastructure. The intuition is that fire hydrants are installed at the same time as their associated water mains, hence the age and type of the hydrant serve as visual indicators of the age of the infrastructure below the surface. Thus, the make and model of each hydrant should provide some indication of the quality of the nearby water infrastructure. We were able to obtain a dataset of all fire hydrants, including their types and addresses, from the city. We used the Google Maps API to match these locations with precise GPS coordinates. When training our models, we include the type of each parcel's nearest fire hydrant as a feature.


\section{Predictive Modeling} \label{sec:model}

\begin{figure}[t!]
\centering
\includegraphics[width=\columnwidth]{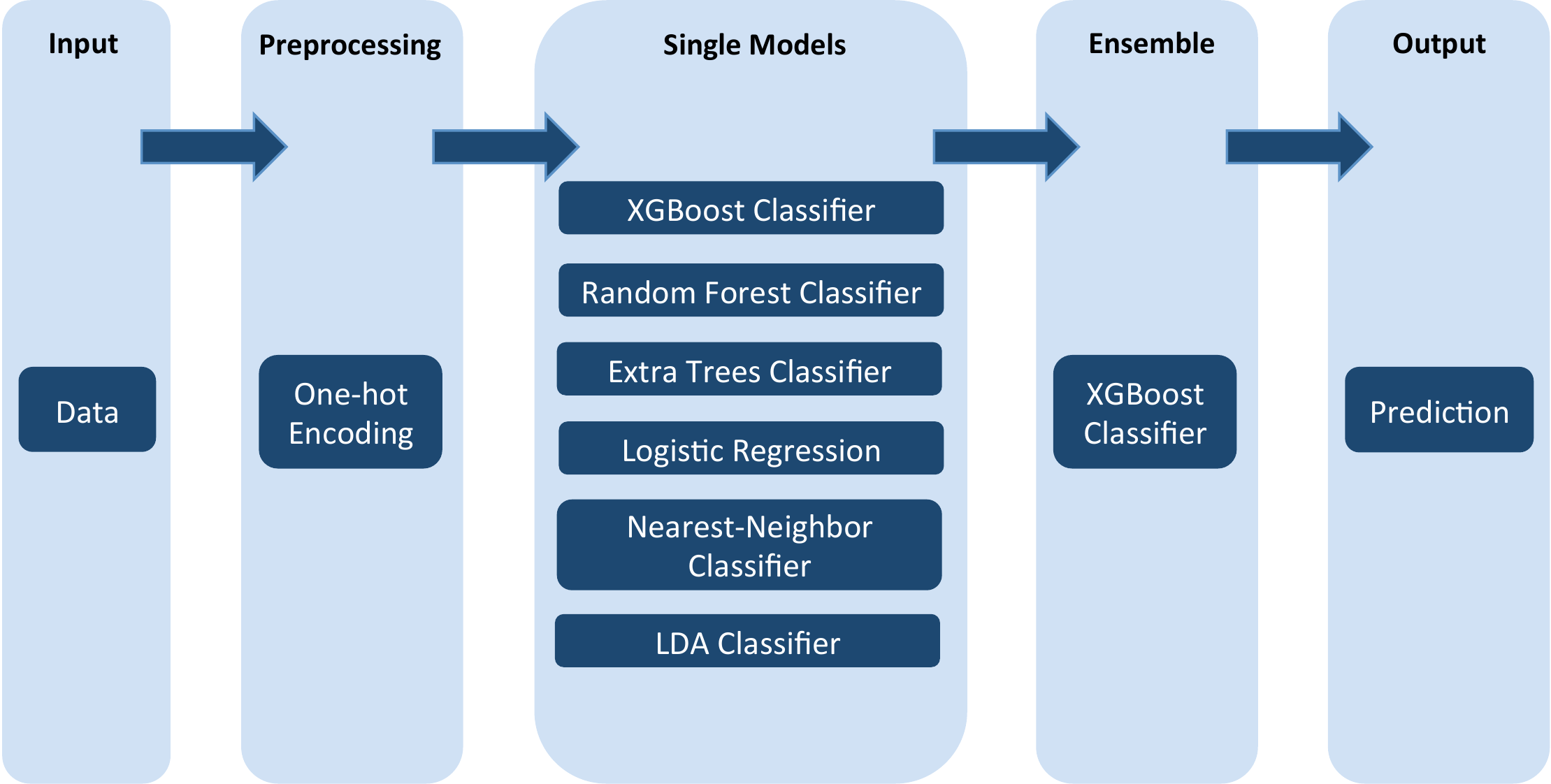}
\caption{
\label{fig:Flow_Chart}
The flowchart of our prediction model.
}
\end{figure}

We develop an ensemble of predictive models to predict whether a given parcel’s lead level will be below or above the EPA and CDC action level of 15 (ppb) of lead in water. Our method has two layers, with the first layer including XGBoost\cite{DBLP:journals/corr/ChenG16}, random forest\cite{breiman2001random}, extremely randomized trees\cite{geurts2006extremely}, logistic regression\cite{Guisan2002}, nearest neighbor\cite{Cover:2006:NNP:2263261.2267456}, and linear discriminant analysis (LDA) classifiers\cite{AHG:AHG2137}, and a second layer of a single XGBoost classifier for ensembling. The flowchart of our prediction model is shown in Figure \ref{fig:Flow_Chart}. The models were trained on 15,447 testing results from 7,999 parcels, and predicted on the other 47,894 parcels in Flint.

\subsection{Feature processing}

In total, we gather 35 features for each sample from the parcel, service line, and fire hydrant dataset. Our models are trained to perform binary classification, where samples with a lead level greater than $15$ (ppb) are considered to be positive and all others are negative. 
A full list of the features and their descriptions are listed in the Appendix \ref{app:datasetDes}. One-hot encoding was performed for categorical features before data were fed to logistic regression, nearest neighbor, and linear discriminant analysis (LDA) classifiers. 

\subsection{Models} 

All the models we use, except XGBoost, were implemented using \texttt{scikit-learn}\cite{scikit-learn}.
For each model, the hyperparameters were determined by a 50-fold cross-validation to minimize the logarithmic loss.
Leave-1-out cross-validation is the preferred method for the cross-validation, due to computational costs we increase the number of folds instead.
It also allows us to take advantage of almost full dataset, each time it exclude only $2\%$ of the data for the training. 
Each time we split the data to training and validation sets, we made sure that data from the same parcels did not exist in both sets to avoid data leakage. Some important hyperparameters are summarized in Table~\ref{tbl:hyperparameters} for each model.

\begin{table}
\centering
\begin{tabular}{l|p{5.5cm}}
\hline
Classifier & Hyperparameters\\
\hline
XGBoost & 200 trees with a maximum depth of 5 \\
Random Forest & 1000 trees with a maximum depth of 9 \\
Extra Trees  & 1000 trees with a maximum depth of 9 \\
KNN & 100 nearest neighbors with manhattan distance (L1) for the Minkowski metric \\
Logistic Regression & L1 regularization \\
LDA & No shrinkage for covariance matrix\\
\hline
\end{tabular}
\caption{
\label{tbl:hyperparameters}
Summary table of hyperparameters for different classifiers.
}
\end{table}

\subsection{Ensemble Model}

The out-of-fold predictions from each model in the first layer were then used as features for the ensemble model in the second layer. This multi-layered fashion of stacked generalization was first introduced by Wolpert\cite{Wolpert1992241}. An XGBoost model with 800 trees and a max depth of 8 was used in the second layer for ensembling to maximize the area under the curve (AUC).

\begin{table}
\centering
\begin{tabular}{l c c}
\hline
Classifier & AUC & LogLoss\\
\hline
XGBoost & 0.660 & 0.274 $\pm$ 0.048 \\
Random Forest & 0.648 & 0.276 $\pm$ 0.047 \\
Extra Trees  & 0.625 & 0.279 $\pm$ 0.047 \\
KNN & 0.621 & 0.296 $\pm$ 0.068 \\
Logistic Regression & 0.641 & 0.280 $\pm$ 0.049 \\
LDA & 0.549 & 0.286 $\pm$ 0.048 \\
\hline
\textbf{Ensemble} & \textbf{0.677} & \textbf{0.273 $\pm$ 0.054} \\
\hline
\end{tabular}
\caption{
\label{tbl:metrics}
Summary table of error metrics for different classifiers.
}
\end{table}

\subsection{Results}

\begin{figure}
\centering
\includegraphics[width=\columnwidth]{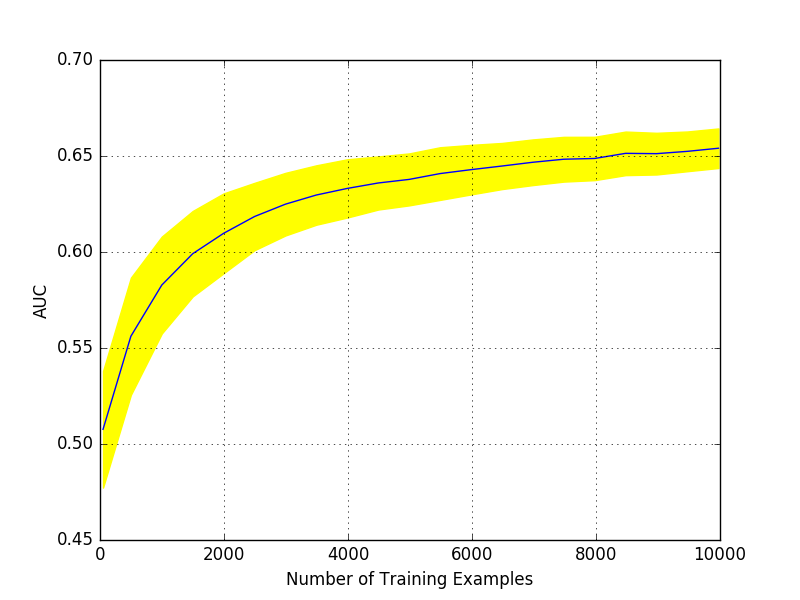}
\caption{
\label{fig:learning_curve}
Learning curve for the XGBoost classifier. We averaged AUC over 400 bootstrap samples, and the highlighted region is showing one standard deviation. }
\end{figure}

\begin{figure}[h]
\centering
\includegraphics[width=\columnwidth]{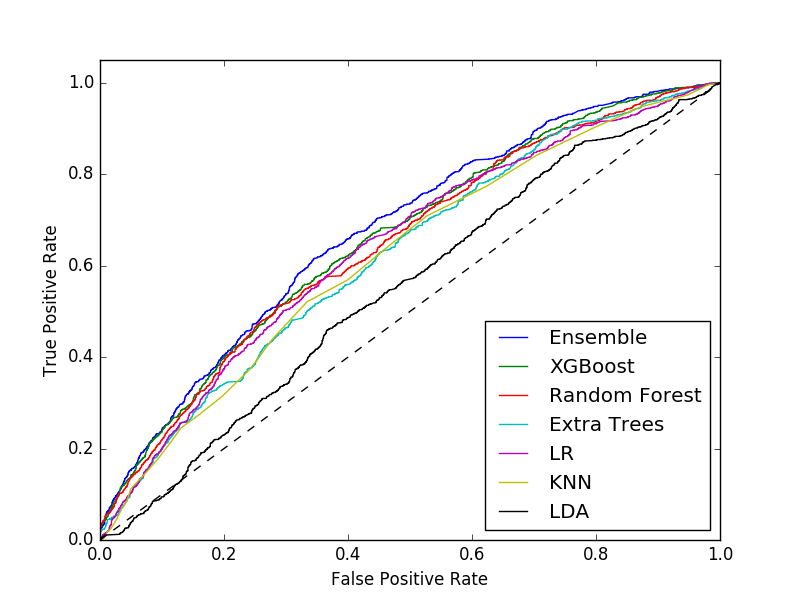}
\caption{
\label{fig:roc}
ROC curves of classifiers adapted in this work. The ensemble model outperforms all individual classifiers.}
\end{figure}

The error metrics including AUC and logarithmic loss are summarized in Table~\ref{tbl:metrics} for each first-layer model as well as the ensemble model. The ensemble model outperforms the XGBoost classifier, which is our best model in the first layer, with AUC of 0.677 and logarithmic loss of 0.273. Much attention has been paid to AUC as we were looking for a classifier that could clearly separate the parcels with over 15 (ppb) of lead in water (positive) from the rest (negative). Our ensemble model also has a higher true positive rate than all the single models at most false positive rates in the receiver operating characteristic (ROC) curves. Figure ~\ref{fig:roc} is showing the ROC curve of each model and ensemble model.
While AUC score improvement for ensemble model is not statistically significant compare to XGBoost, ensemble models are good in decreasing the bias in the error, which allows more accurate probability calibration.  

Figure~\ref{fig:learning_curve} shows the learning curve for the XGBoost classifier, which is our best model in the first layer. Here we reserve 5,503 examples for validation and use different subsets of the remaining 9,974 examples for training. Similar to the cross validation, we made sure that the same parcels do not exist in both training and validation sets to avoid data leakage. The AUC score increases as more training examples are included, and there seems to be room for improvement beyond 10,000 training examples.

\begin{figure}[h]
\centering
\includegraphics[width=0.8\columnwidth]{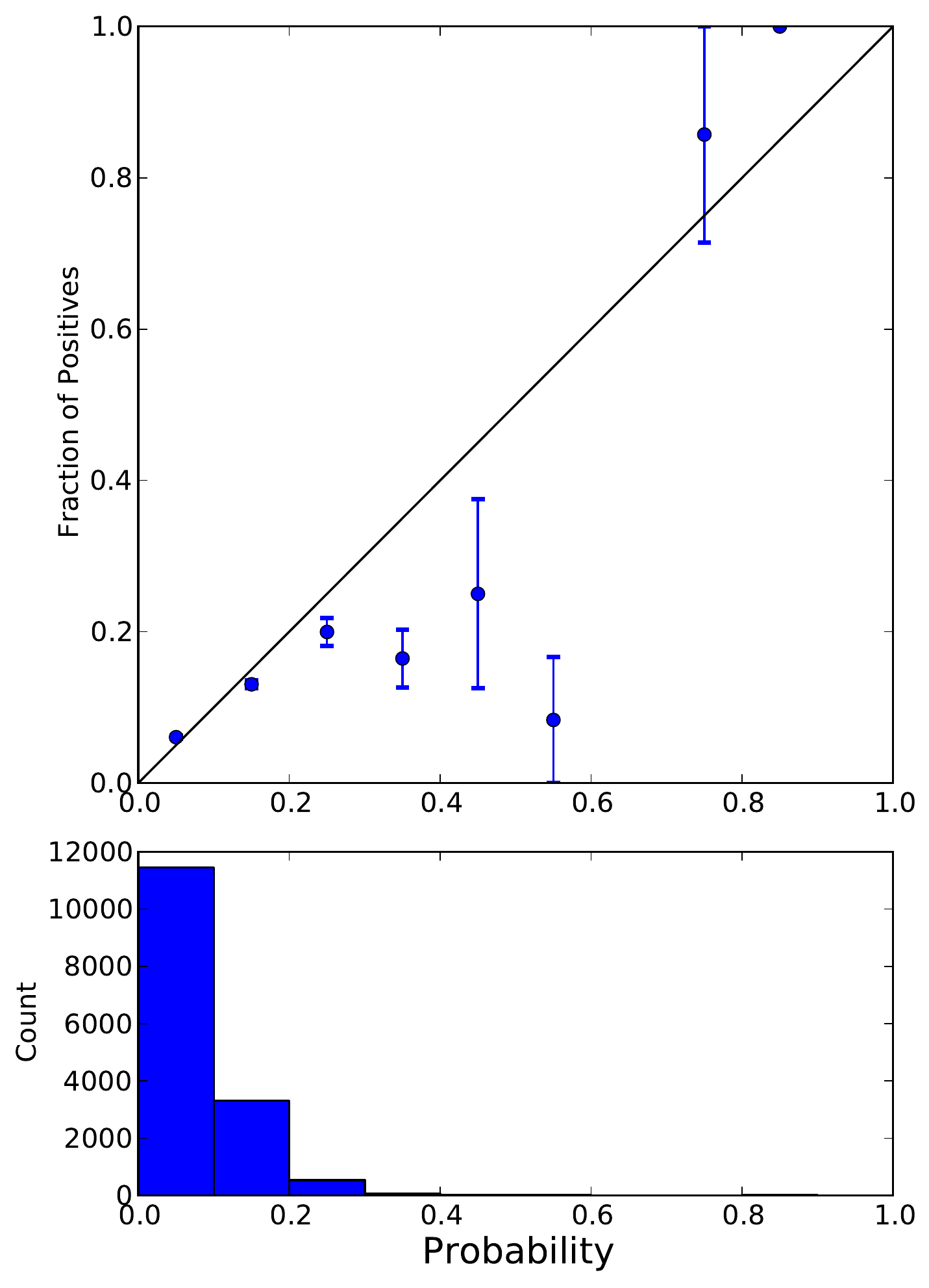}
\caption{
\label{fig:calibration}
Calibration curve of the ensemble classifier. The error bars are calculated by bootstrapping parcels in each probability bin.  The bottom panel shows the number of samples in each probability bin.}
\end{figure}

While we focus on the separation of positive and negative instances, we also want to make sure that the predicted probability of positive label for each instance is calibrated properly. Ideally the probabilities predicted by a well-calibrated classifier can be directly interpreted as the fraction of positives in the dataset predicted to have similar probabilities of positive labels. The ensemble model is found to be well calibrated at small predicted probabilities but deviates considerably from the perfectly calibrated case at probabilities larger than 0.3. Figure~\ref{fig:calibration} shows the fraction of parcels above EPA action level in predicted probability bins. This is largely attributed to the fact that the positive and negative classes are very imbalanced in the training data, and only 8.3\% of the testing results are above the EPA and CDC action level of 15 (ppb) of lead in water (positive).



\section{Service Lines, Property Age, and Other Risk Factors}

In addition to building a predictive model for lead levels, we aim to identify specific features that are strong predictors of high lead levels. These features will form our most important set of risk factors, and in this section we refer to features and factors interchangeably.


Knowing the risk factors for any particular parcel allows us to quickly identify whether it carries a high risk of having lead above the EPA action level. 
A predictive set of risk factors can thus provide officials with an efficient way of quickly identifying at-risk areas. 
We are cautious not to use the term ``causal factors" in this context. 
One should note that the identified risk factors are not necessarily the causes of lead contamination in the water. Rather, these features are those which allow us to separate potential parcels with a high risk of having unsafe lead levels according to our classification.

\subsection{Service Lines and Year of Construction}

\begin{figure}[h]
\centering
\includegraphics[width=\columnwidth]{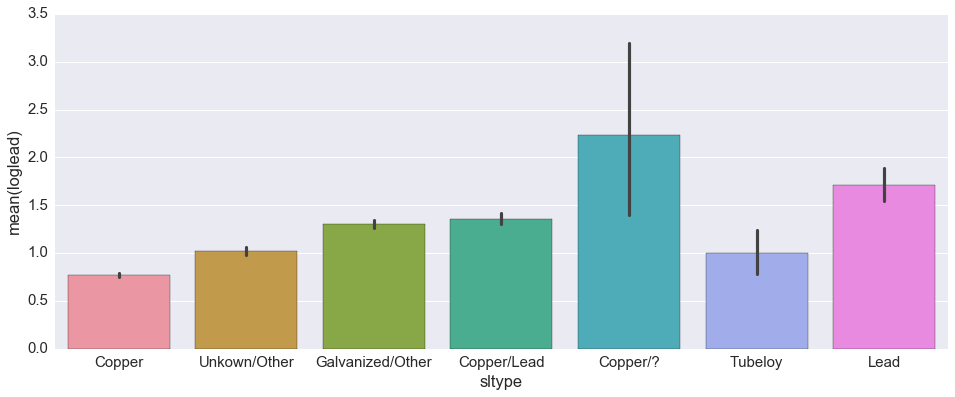}
\caption{
\label{fig:lead_by_sltype}
Lead levels by Service Line Type. We see a statistically significant difference in the mean for homes whose city records report copper, versus homes with records reporting lead.}
\end{figure}

Much of the media focus after news of the water crisis broke was centered around the problem of Flint's service lines\footnote{http://michiganradio.org/post/flint-mayor-city-will-remove-lead-service-lines-high-risk-homes}. As all water entering the home must pass through the service line, these pipes make an easy culprit as to the source of the lead. If we use the city records as a rough estimate of which homes possess lines with lead material, we can look at average (log) lead levels over all homes that submitted a residential water test for which we have a record. We report the mean of $\log(1 + \text{Lead\_in\_ppb})$ for all of these water tests in Figure~\ref{fig:lead_by_sltype}.  We recall that the city records are quite noisy, as we discussed in Section~\ref{sec:serviceLines}.

\begin{figure}[h]
\centering
\includegraphics[width=\columnwidth]{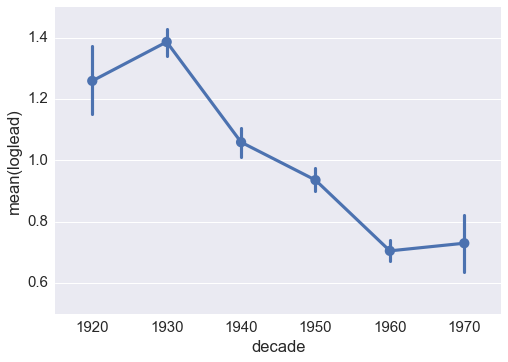}
\caption{
\label{fig:lead_by_year}
The average $\log(1 + \text{Lead\_in\_ppb})$ for residential water readings for various decades of property construction, ranging from 1920-1979, a period during which over 85\% of current homes in Flint were built.}
\end{figure}

\begin{figure}[h]
\centering
\includegraphics[width=\columnwidth]{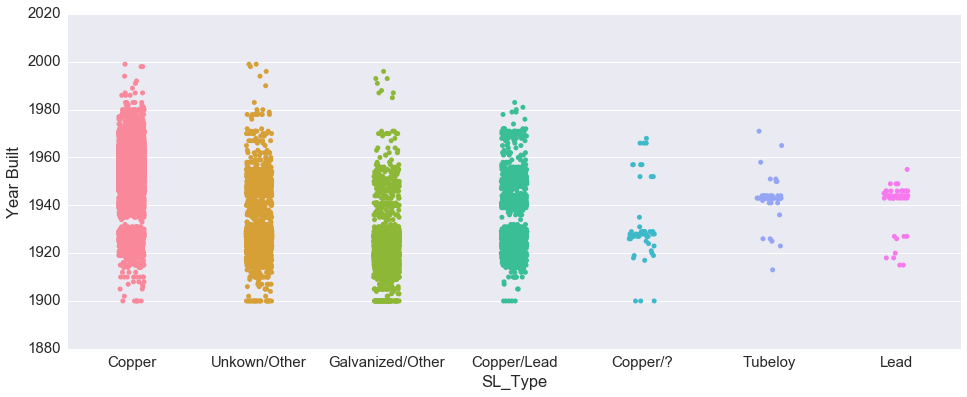}
\caption{
\label{fig:sltype_year}
When were different service line types used in Flint, according to city records. Every dot represents a given property for which an SL record exists.}
\end{figure}

Given the clear statistically significant difference between lead levels for homes with copper versus lead service lines, it would be easy to draw the conclusion that the service line is the primary driver of lead in the water. But we would cast some doubt on this simple narrative, as one can find other aspects of the various properties in Flint that have high correlation with the lead levels. Interestingly, one observes that the \emph{property age} is strongly associated with elevated lead, with a significant drop between the 1930s and the 1960s. We give a plot of average lead levels by decade in Figure~\ref{fig:lead_by_year}. We also show various service line types were used at different periods in Figure~\ref{fig:sltype_year}.

We still do observe elevated lead levels for many homes constructed in the 1960s and 1970s, and during this period it was very rare to use lead piping. On the other hand, during these years it was still possible to purchase fixtures that contained lead or lead alloys, and thus home faucets could be the source of lead contamination. The use of lead solder, as well as lead pipes, was not banned until 1986 \cite{calabrese1989safe}. Unfortunately we do not have data on the use of solder and fixture types within homes which is a challenge in assessing their level of contribution to contaminated drinking water.

\subsection{Risk Factors via Predictive Modelling}

In the remainder of this section, we consider the importance of various risk factors by way of their predictive power in determining elevated lead. We now identify the 10 most predictive factors in the risk assessment analysis using our first-layer XGBoost model. This is done because its metrics scores are close to those of the ensemble model, and because it is the most predictive model of the first layer.
We drop one feature at a time from the predictive model, ranking the importance of each feature based on the corresponding drop in AUC score. These conclusions remain the same when performing the same analysis for the Random Forest and Extra Trees models.
Table \ref{tab:importantFeature} shows the 10 most predictive factors of XGBoost based on the drop in AUC metric after removing them. 

\subsubsection*{Local factors}

XGBoost picks up geographic features, such as Longitude, Latitude, and PID, as belonging to the 10 most predictive variables. We note that the unique parcel identifiers (PIDs) are geographically determined and are related to the zip code of the corresponding parcel. 
This indicates that geographic location is one of the most important predictive features. This could indicate that there are problem hotspots where local pipelines are affected disproportionately more than other areas, or that contaminated houses can effect their neighbors.
There also may be other local features which are not captured in our dataset influencing their importance.
While causal factors are uncertain, we are predicting neighborhoods which are more likely to be at risk of lead contamination.

\subsubsection*{Property features}

The property features ``Land Value" and ``HomeSEV" also appear in the top 10 most predictive risk factors.
This could be due to the clustering of houses in one area that may contribute to the lead in the water. 
For example, some old houses have lead in their pipelines or fixtures which may contribute to the lead contamination \footnote{ Note that, in this work, we do not attempt to isolate the contribution of local environment causal factors from properties itself causal factors. }.

\subsubsection*{Service Lines}

As expected, the type of service line and whether the service lines are made out of lead are important factors in determining the risk level. 
Table \ref{tab:importantFeature} indicates that local and property factors are far more important in lead contamination prediction than the service line alone. 
Our analysis shows that notable number of homes with non-lead service lines are experiencing high lead contamination.
It can be partially due to the mis-classification of service lines, as discussed in section \ref{sec:serviceLines}, and most importantly due to other potential causal factors, such as age of house.
There also could be unmeasured factors, like in house plumbing or age of water in the main pipelines.

\begin{table}
\centering
\begin{tabular}{l c}
\hline
Rank & XGBoost\\\hline
1  & Longitude \\
2  & PID \\
3  & SL Type \\
4  & Owner Type  \\
5  & Property Zip Code \\
6  & PRECINCT \\
7  & HomeSEV  \\
8  & Hydrant Type \\
9  & SL Type2 \\
10  & Land Improvements Value\\
\hline
\end{tabular}
\caption{Summary table of 10 most predictive risk factor for XGBoost.}\label{tab:importantFeature}
\end{table}

\subsection{Neighborhood Risk Assessment}

\begin{figure}[h]
\centering
\includegraphics[width=.5\textwidth]{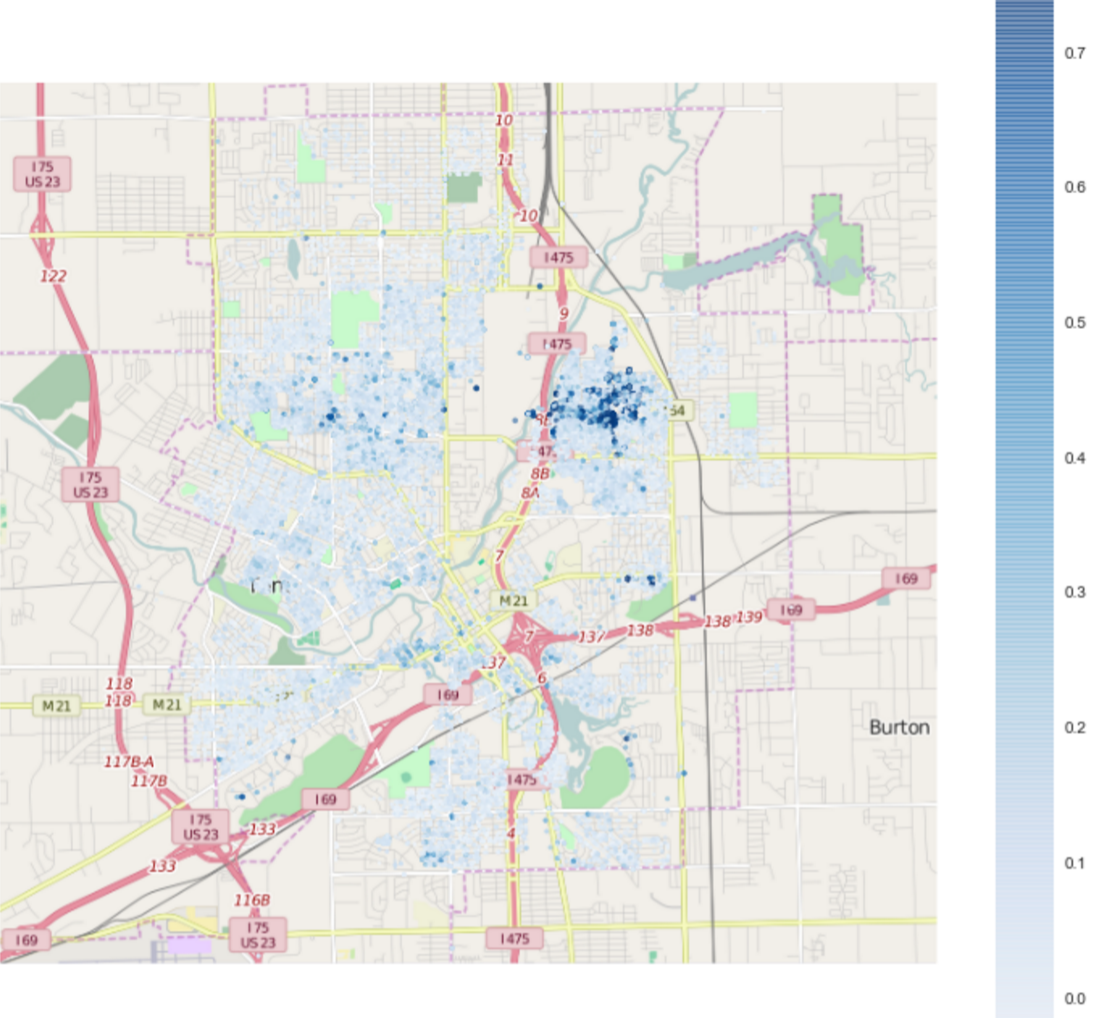}
\caption{Parcels at risk of lead contamination as predicted by our ensemble model. Color corresponds to the predicted probability of lead contamination above 15 (ppb). Only parcels with a predicted risk $0.1$ are pictured.}
\label{fig:prediction}
\end{figure}

We use the prediction model, described in section \ref{sec:model}, to predict whether houses, which did not submit any samples, are above EPA action level or not.
The model allows us to predict the probability of lead contamination above EPA action level for individual homes.
Figure \ref{fig:prediction} is showing parcels at risk of lead contamination as predicted by our ensemble model. Color corresponds to the predicted probability of lead contamination above 15 (ppb) for individual parcels. Only parcels with a predicted risk greater than $0.1$ are visualized.
Though there is variation in the map, but it appears that there are clusters of neighborhoods which are potentially at high risk of lead exposure due to their water quality. 
Identifying these neighborhood allows policy makers plan accordingly and set their priorities.

\section{Conclusion}

The lead contaminating Flint's water systems poses a serious health risk for all of the city's residents and those in surrounding areas. We collaborated with the City of Flint and the Michigan Department of Environmental Quality to collect data.

Using this data, we constructed a model to predict which locations are most likely to have water with lead contamination above the EPA action level of 15 PPB. In working to identify which features are strong predictors of high lead levels, we found that a number of factors, not just the composition of service lines, are important to consider in addressing the ongoing crisis. Knowing these risk factors can help policy makers and community members better allocate limited resources and prioritize action in this time of need.

\section*{Acknowledgments}
Much of this work was supported by the National Science Foundation under CAREER grant IIS-1453304, an award that helped facilitate the development of the Michigan Data Science Team. We also acknowledge support from {Google.org}, who provided a \$150,000 grant to the University of Michigan (Flint and Ann Arbor campuses) for the development of a citizen-focused mobile app for Flint residents -- this award initiated our heavy focus on data science tools to aid the water issues in Flint. Finally we would like to thank Advanced Research Computing at the University of Michigan for graciously providing computing resources via the Flux platform.

\nocite{*}
\bibliographystyle{abbrv}
\bibliography{references}

\appendix

\section{Description of Dataset columns}  \label{app:datasetDes}
 
A summary of the training dataset columns is given in following.

\small{
\texttt{Lead (ppb)} - Lead level in the submitted sample (ppb). \hfill\break
\texttt{PID} - Unique parcel ID.\hfill\break
\texttt{Property Zip Code} - property zip code \hfill\break
\texttt{Owner Type} - owner type: including residential, commercial, and industrial \hfill\break
\texttt{Homestead} - Homestead is a person's or family's residence, which comprises the land, house, and outbuildings, and in most states is exempt from forced sale for collection of debt. \hfill\break 
\texttt{Homestead Percent} - 0-100 \hfill\break
\texttt{HomeSEV} - SEV is State Equialized Value. That’s what the government thinks your home is worth.\hfill\break
\texttt{Land Value} - Land value \hfill\break
\texttt{Land Improvements Value} - Value of improvements on the parcel \hfill\break
\texttt{Residential Building Value} - Residential building value (only for residential buildings) \hfill\break
\texttt{Commercial Building Value} - Commercial building value (only for commercial buildings) \hfill\break
\texttt{Building Storeys} - Number of storeys \hfill\break
\texttt{Parcel Acres} - Parcel acres \hfill\break
\texttt{Use Type} - Residential, commercial or industrial use. \hfill\break
\texttt{Prop Class} - Whether a parcel is agricultural, industrial, residential, or commercial property. \hfill\break
\texttt{Old Prop class} - Previous Prop Class \hfill\break
\texttt{Year Built} - Year which the building is built \hfill\break
\texttt{USPS Vacancy} - Vacancy status of property according to USPS records \hfill\break
\texttt{Zoning} - City of Flint zoning assignment. \hfill\break
\texttt{Future Landuse} - Planned use for land in the future \hfill\break
\texttt{DRAFT Zone} - Future assigned zoning \hfill\break
\texttt{Housing Condition 2012} - Building condition according to the city record in 2012 (only for residential properties) \hfill\break
\texttt{Housing Condition 2014} - Building condition according to the city record in 2014 (only for residential properties) \hfill\break
\texttt{Commercial Condition 2013} - Building condition according to the city record in 2013 (only for residential properties) \hfill\break
\texttt{Rental} - Rental Residential Building or not \hfill\break
\texttt{Residential Building Style} - Style of Residential Building \hfill\break
\texttt{Latitude, Longitude} - Latitude and Longitude \hfill\break
\texttt{Hydrant Type} - Type of closet hydrant to the property \hfill\break 
\texttt{Ward} - A ward is an optional division of a city or town for  administrative and representative purposes, especially for purposes of an election. \hfill\break
\texttt{PRECINCT} - Voting location the parcel belongs to. \hfill\break
\texttt{CENTRACT} - A census tract/area is geographic region defined for the purpose of taking a census. Numbers in the column are the population sizes (number of people). \hfill\break
\texttt{CENBLOCK} - A census block is the smallest geographic unit used by the United States Census Bureau. \hfill\break
\texttt{SL\_Type} - Service line connection type. \hfill\break
\texttt{SL\_Type2} - Second service line connection type, if more than one connection for one parcel. \hfill\break
\texttt{SL\_Lead} - Lead/No Lead connection
}

\end{document}